\documentclass{article}
\usepackage[utf8]{inputenc}
\usepackage{authblk}
\usepackage{setspace}
\usepackage[margin=1.25in]{geometry}
\usepackage{graphicx}
\graphicspath{ {./figures/} }
\usepackage{subcaption}
\usepackage{amsmath}
\usepackage{amssymb}
\newcommand{\argmax}{\mathop{\rm argmax}\limits}
\usepackage{hyperref}
\usepackage{algorithm}
\usepackage{algorithmic}
\usepackage{orcidlink}

\hypersetup{
    colorlinks,
    citecolor=blue,
    filecolor=blue,
    linkcolor=blue,
    urlcolor=blue
}

\usepackage[style=ieee, 
citestyle=numeric-comp,
sorting=none]{biblatex}
\makeatletter
\renewcommand{\maketitle}{\bgroup\setlength{\parindent}{0pt}
\begin{flushleft}
  {\large\textbf{\@title}}

  \@author
\end{flushleft}\egroup
}
\makeatother

\title{Bandit approach to conflict-free multi-agent Q-learning in view of photonic implementation}
\author[1, *]{Hiroaki Shinkawa\,\orcidlink{0000-0002-2999-1413}}
\author[1]{Nicolas Chauvet\,\orcidlink{0000-0002-6504-1730}}
\author[1]{Andr\'{e} R\"{o}hm\,\orcidlink{0000-0002-8552-7922}}
\author[1]{Takatomo Mihana\,\orcidlink{0000-0002-4390-710X}}
\author[1]{Ryoichi Horisaki\,\orcidlink{0000-0002-2280-5921}}
\author[2]{Guillaume Bachelier\,\orcidlink{0000-0002-7990-7857}}
\author[1]{Makoto Naruse\,\orcidlink{0000-0001-8982-9824}}

\affil[1]{Department of Information Physics and Computing, Graduate School of Information Science and Technology, The University of Tokyo, 7-3-1 Hongo, Bunkyo-ku, Tokyo 113-8656, Japan.}
\affil[2]{Univ. Grenoble Alpes, CNRS, Institut N\'{e}el, 38000 Grenoble, France.}
\affil[*]{Corresponding author. Email: gokukyukyoku555@gmail.com}

\date{}

\begin{document}

\maketitle

\begin{abstract}
Recently, extensive studies on photonic reinforcement learning to accelerate the process of calculation by exploiting the physical nature of light have been conducted.
Previous studies utilized quantum interference of photons to achieve collective decision-making without choice conflicts when solving the competitive multi-armed bandit problem, a fundamental example of reinforcement learning.
However, the bandit problem deals with a static environment where the agent's action does not influence the reward probabilities.
This study aims to extend the conventional approach to a more general multi-agent reinforcement learning targeting the grid world problem.
Unlike the conventional approach, the proposed scheme deals with a dynamic environment where the reward changes because of agents' actions.
A successful photonic reinforcement learning scheme requires both a photonic system that contributes to the quality of learning and a suitable algorithm.
This study proposes a novel learning algorithm, discontinuous bandit Q-learning, in view of a potential photonic implementation.
Here, state-action pairs in the environment are regarded as slot machines in the context of the bandit problem and an updated amount of Q-value is regarded as the reward of the bandit problem. 
We perform numerical simulations to validate the effectiveness of the bandit algorithm.
In addition, we propose a multi-agent architecture in which agents are indirectly connected through quantum interference of light and quantum principles ensure the conflict-free property of state-action pair selections among agents. 
We demonstrate that multi-agent reinforcement learning can be accelerated owing to conflict avoidance among multiple agents.
\end{abstract}


\section{Introduction}
Reinforcement learning is a machine learning technique that enables an agent to perform the desired task through repeated trials and errors in a particular environment \cite{sutton2018reinforcement}.
Methods implemented in previous studies have yielded remarkable results, including victories over professional human players in board games, such as Go \cite{silver2016mastering, silver2018general}.
Recently, photonic approaches to reinforcement learning to outsource the computational costs and exploit the physical nature of light have been proposed \cite{flamini2020photonic, steinbrecher2019quantum, saggio2021experimental, bukov2018reinforcement, bueno2018reinforcement}.

Previous studies solved the bandit problem, a fundamental reinforcement learning model, using the quantum nature of photons \cite{naruse2015single, chauvet2019entangled, chauvet2020entangled, amakasu2021conflict}.
The bandit problem is a frequently used model of human decision-making \cite{daw2006cortical}.
Multiple slot machines probabilistically generate a reward and an agent attempts to maximize the cumulative reward from the machines under the constraint that they can only play one machine at a time \cite{sutton2018reinforcement, maghsudi2016multi}.
Because the agent lacks prior knowledge of the reward probabilities of the machines, they must play with various machines, including bad machines at that time, in the early stages of the game to accurately estimate the reward probabilities.
This results from the stochastic nature of reward generation; that is, a machine should not be considered as having a low reward probability just because it has not been generating many rewards at that time.
However, the agent would suffer a loss if they played bad machines excessively; therefore, they must concentrate on the machines that have the highest reward probabilities in the latter stages of the game.
The former aspect is called exploration, whereas the latter is called exploitation; balancing these two conflicting demands is the key to solving this problem \cite{march1991exploration}.
The softmax rule is a model that balances exploration and exploitation through probabilistic decision-making and is considered as the model that best fits human decision-making \cite{daw2006cortical}.

The quantum nature of photons can be applied to solve the bandit problem.
In particular, by mapping the selection of a machine to the observation of a photon's state, probabilistic decision-making can be implemented because the state observed is determined probabilistically \cite{naruse2015single}.
Furthermore, the role of photons in decision-making becomes critical owing to entanglement and quantum interference, which are inherent properties of quantum physics \cite{chauvet2019entangled, chauvet2020entangled, amakasu2021conflict}.
For example, consider a situation in which two agents solve the bandit problem simultaneously but the selection of the same machine reduces the total reward.
This is analogous to a real-world situation when multiple people or devices simultaneously connect to the same wireless channel, resulting in the degradation of the individual communication speed \cite{lai2010cognitive, kim2016harnessing, besson2018multi,maghsudi2016multi}.
By observing the states of a two-photon pair whose polarizations are entangled, the two agents can ensure that their choices always differ in such circumstances. 
That is, entanglement avoids selection conflicts.

Chauvet {\it{et al.}} theoretically and experimentally showed that the competitive multi-armed bandit problem, which deals with the aforementioned situation, can be resolved with no conflict of choices by two agents faced with two machines using photon pairs whose polarizations are entangled \cite{chauvet2019entangled, chauvet2020entangled}.
Their system is remarkable in that the agents can avoid selection conflicts without directly communicating with each other about the machine to be selected because of quantum entanglement.
Furthermore, Amakasu {\it{et al.}} theoretically showed that the system could be extended to handle three or more machines using quantum interference of orbital angular momentum of light \cite{amakasu2021conflict}.
Accordingly, they developed a photonic system that ensured conflict-free selections by two agents with an arbitrary number of machines.
In addition, Shinkawa {\it{et al.}} formulated a problem in which people individually have a probabilistic preference over options, derived the optimal joint decision-making in terms of satisfaction \cite{shinkawa2022optimal}, and demonstrated that a system based on quantum interference of photons can provide a heuristic solution to this problem \cite{shinkawa2022conflictfree}. This is another example of the coordination of multi-individual decision-making by a photonic system.

This study aims to demonstrate the potential of a photonic reinforcement learning scheme, which requires the combination of a suitable algorithm and a photonic system that leverages the unique physical nature of photons.
Based on previous studies, a photonic system can be used to solve the bandit problem, a simple reinforcement learning task.
However, to tackle challenging problems, the photonic system must be extended such that it can handle three or more agents, and the algorithm must be modified accordingly.
The environment in the bandit problem is static, whereas that in a general reinforcement learning problem is generally dynamic.
In particular, the environment (reward probabilities) is independent of the agent's action in the bandit problem. 
Conversely, in a general reinforcement learning problem, the state of the environment changes because of the action, which must be considered in the learning process.
This study presents a modified algorithm that can solve a broader class of reinforcement learning problems.
While the extension of the photonic system with more than two agents remains open and must be addressed in future studies, this study lays the foundation for a photonic reinforcement learning scheme that can be implemented once the photonic system is developed.

We consider the grid world problem as a dynamic environment \cite{sutton1990integrated}.
It is a collection of cells in which an agent can either implement an up, down, left, or right action.
Depending on the combination of cells and actions, the agent receives different rewards from the environment.
Because different cells have different reward environments, the grid world is a dynamic environment.

While Q-learning is generally used as an algorithm for reinforcement learning \cite{watkins1989learning, watkins1992q, jang2019q}, this study proposed a combination of Q-learning with the bandit algorithm, called discontinuous bandit Q-learning (DBQL).
Although Q-learning aims to learn the optimal paths, this study aims to learn the value of each state-action pair in the entire environment with high accuracy.
Thus, suppose the agent deviates from the optimal paths. In that case, it will accurately return to the optimal paths from any location in the environment.

In the proposed DBQL method, each agent selects a state-action pair in the environment at each time step and updates the corresponding Q-value (a detailed definition is given in Sec. \ref{sec:methods}).
Decisions on the state-action pair to be selected have a similar structure to the bandit problem because the agent must balance two demands; the first is the demand for exploitation, which is to update state-action pairs that are likely to have a large value of $\Delta Q$ (the change in the Q-value) for the moment, to accelerate learning. The second is the demand for exploration to accurately estimate the expected value of $\Delta Q$ for other state-action pairs that have not yet been visited frequently.
Thus, by considering the state-action pairs as machines and $\Delta Q$ as a reward, the accurate estimation of Q-values for the entire environment can be viewed as a bandit problem.

Furthermore, we consider a case in which multiple agents participate in the learning and follow DBQL simultaneously.
We demonstrate the learning can be accelerated by avoiding the selection of the same state-action pair simultaneously; that is, by forcing the agents to make conflict-free decisions.
As earlier mentioned, we have not conceived a photonic system that enables conflict-free selections among more than two agents without direct communication yet.
Accordingly, this study algorithmically realized conflict-free selections, which essentially means forcing the agents to disclose their selections.
Once a photonic system with more than two agents is developed in the future, our scheme will be implemented by the mixture of the photonic system and our proposed algorithm, thus eliminating the necessity for the agents to share their selections.

The remainder of this paper is organized as follows.
Sec. \ref{subsec:grid_world} describes the experimental environment and the grid world. 
Section \ref{subsec:dbql} explains Q-learning followed by a detailed description of the proposed method DBQL. 
Section \ref{subsec:multi_agent} provides the environment's response when multiple agents explore simultaneously, and Sec. \ref{subsec:avoidance} illustrates a selection conflict avoidance system using quantum interference of photons.
Section \ref{sec:result} demonstrates the result of performing an actual search in the grid world using DBQL to quantify the impact of the bandit algorithm on learning and the impact of avoiding selection conflicts.
Finally, Sec. \ref{sec:discussion} discusses the results and future perspectives.

\section{Materials and Methods}\label{sec:methods}
\subsection{Experimental Design}\label{subsec:grid_world}
\begin{figure}[htp]
    \centering
    \includegraphics[width=0.5\textwidth]{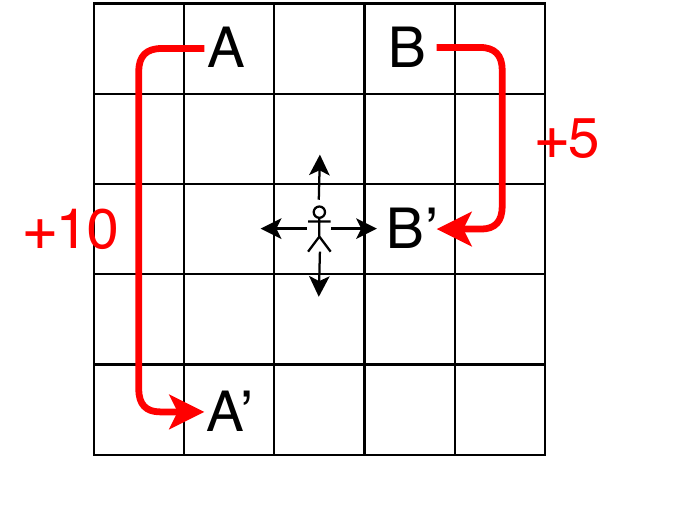}
    \caption{$5 \times 5$ grid world. The agent implements one of the four actions at each time step and receives a reward and the next state. In special cells A and B, the reward is large and the agent jumps to another cell.}
    \label{fig:grid_world}  
\end{figure}

The schematic of the grid world, which is often used as a model in previous studies on reinforcement learning \cite{sutton1990integrated} is shown in Fig. \ref{fig:grid_world}.
An agent exists in the grid world and moves around in the environment.

In this example, the world is represented by a $5 \times 5$ cell grid, where each cell is called a ``state.''
At each time step, the agent selects an ``action'' either up, down, left, or right.
In the grid world, when an agent is in a state $s_t$ at a time step $t$, the chosen action $a_t$ determines the reward $r_t$ and the next state $s_{t+1}$, which is provided by the environment.
In this study, we assumed the environment is Markovian, meaning the next state $s_{t+1}$ is determined only by the current state $s_t$ and action taken $a_t$.
For example, if an agent is in the top left corner cell and selects the action ``right,'' the agent earns a specific reward and moves to cell A.
Herein, the rule that determines the action to be chosen by the agent in each state is called a ``policy.''
In this study, we confine the policy to be deterministic.

Then, the ``action-value function'' $Q^\pi (s, a)$ is determined for each state-action pair $(s, a)$ when the agent follows a particular policy $\pi$.
This function represents the total future rewards when the agent is currently in a state $s$ and takes an action $a$ followed by a series of actions that $\pi$ instructs.
\begin{equation}
    Q^\pi(s, a) = \sum_{t=0}^\infty \gamma^t r_t,
\end{equation}
where $\gamma$ is the time discount.
The time discount is applied to reflect that distant-future rewards matter less than near-future rewards and to ensure the convergence of the function.
Note that a larger $\gamma$ means the agent is more concerned about a long-term benefit.
If the reward is determined stochastically, the expected value of the reward $\mathbb{E}[r_t]$ is used instead of $r_t$.

Suppose the grid world problem is fully known, meaning all the possible states, actions, and rewards are known in advance.
In that case, we can use dynamic programming algorithms, such as value iteration or policy iteration, which solve the Bellman equation to derive the optimal action-value function $\tilde{Q}(s, a)$ and policy $\tilde{\pi}(s)$, where
\begin{equation}
    \tilde{\pi}(s) = \argmax\limits_a\tilde{Q}(s, a)    
\end{equation}
is satisfied \cite{bellman1966dynamic, bellman1954theory}.
This study aims to ensure that the agent without the knowledge of the environment accurately learns the optimal action-value function $\tilde{Q}(s, a)$ for all state-action pairs $(s, a)$ using the information they obtain from the environment.
The initial values are $Q(s, a) = 0$, and the value of $Q(s,a)$ during the learning is called ``Q-value.''

\subsection{Discontinuous Bandit Q-Learning}\label{subsec:dbql}
Q-learning is generally used to solve the grid world. 
Algorithm \ref{alg:ql} provides an overview of Q-learning \cite{watkins1992q}.

\begin{algorithm}[H]
    \caption{Q-learning}
    \label{alg:ql}
    \begin{algorithmic}[1]
    \STATE Pick an initial state $s_0$
    \WHILE{$t \leq T$}
        \IF{rand() $< \epsilon$}
        \STATE $a_t \leftarrow$ random
        \ELSE
        \STATE $a_t \leftarrow \argmax_{a} Q(s_t,a)$
        \ENDIF
        \STATE Implement an action $a_t$ and obtain a reward $r_t$ and the next state $s_{t+1}$. Then, update $Q(s_t,a_t)$:
        \STATE $\quad Q(s_t,a_t) \leftarrow Q(s_t,a_t) + \alpha \cdot \left\{r_t + \gamma \max\limits_{a'} Q(s_{t+1}, a') - Q(s_t,a_t) \right\}$
    \ENDWHILE
    \end{algorithmic}
\end{algorithm}

First, the agent randomly chooses the initial state $s_0$.
The policy $\pi$ is the $\epsilon$-greedy method.
That is, the agent usually chooses an action with the largest $Q(s_t, a_t)$; however, the action is chosen uniformly at random with a probability $\epsilon$.
Notably, the possible actions are confined to the four actions in the current cell, unlike in discontinuous Q-learning proposed later.
Thus, the agent receives a reward $r_t$ and the next state $s_{t+1}$ from the environment after executing the action $a_t$.
Finally, it updates $Q(s_t, a_t)$ according to the following rule:
\begin{equation}\label{eq:q_update}
    Q(s_t, a_t) \leftarrow Q(s_t, a_t) + \alpha \cdot \left\{r_t + \gamma \max\limits_{a'} Q(s_{t+1}, a') - Q(s_t, a_t)\right\},
\end{equation}
where $\alpha$ is the learning rate and $\gamma$ is the time discount.
Note that if $\alpha = 0$, nothing is learned; however, if $\alpha$ = 1, all the previous experiences are forgotten and only the last experience is considered.
This process can be repeated to obtain Q-values as good approximations of the optimal action-value functions $\tilde{Q}(s, a)$.

In this study, we first propose discontinuous Q-learning to interpret the original Q-learning as a decision-making problem about what state-action pair $(s, a)$ the agent selects at each time step.
Algorithm \ref{alg:dql} provides an overview of discontinuous Q-learning.
Unlike basic Q-learning, in discontinuous Q-learning, the new state $s'$ obtained from the environment because of the action $a_t$ is ignored and a new state-action pair $(s_{t+1}, a_{t+1})$ is chosen from all the possible pairs in the environment, which are not limited to pairs whose states are $s'$.
Although the algorithm implies that the agent ``jumps'' at every time step, this assumption is realistic in reinforcement learning.
The initial position is already often randomly chosen in the existing algorithms at the start of every iteration (thus, the position ``jumps'' from the final position in the last iteration) in famous problems such as the cart-pole problem \cite{barto1983neuronlike} or the maze-solving problem \cite{samuel2000some}.

\begin{algorithm}[H]
    \caption{Discontinuous Q-learning}
    \label{alg:dql}
    \begin{algorithmic}[1]
    \WHILE{$t \leq T$}
        \STATE Select one state-action pair $(s_t,a_t)$ based on specific criteria
        \STATE Implement an action $a_t$ and obtain a reward $r_t$ and next state $s'$. Then, update $Q(s_t,a_t)$:
        \STATE $\quad Q(s_t,a_t) \leftarrow Q(s_t,a_t) + \alpha \cdot \left\{r_t + \gamma \max\limits_{a'} Q(s', a') - Q(s_t,a_t) \right\}$
    \ENDWHILE
    \end{algorithmic}
\end{algorithm}

Algorithm \ref{alg:dql} shows that the state-action pair $(s_t, a_t)$ updated by the agent at every time step is determined based on ``specific criteria.''
This study demonstrates that the bandit algorithm can function effectively as the selection criterion.

Now, $\Delta Q(s, a)$ is defined as the absolute value of the updated amount of $Q(s, a)$ at each time step:
\begin{equation}
    \Delta Q(s, a) := \left|\alpha \cdot \left\{r_t + \gamma \max\limits_{a'} Q(s', a') - Q(s_t,a_t) \right\}\right|.
\end{equation}
Larger $\Delta Q(s, a)$ means faster learning.
Therefore, the agent should choose a state-action pair $(s, a)$ with a high expected value of $\Delta Q(s, a)$ to make the learning process more efficient.
However, the potential update $\Delta Q(s, a)$ for other state-action pairs $(s, a)$ could be higher and will also vary as the update proceeds. 
Thus, the agent cannot rely on just selecting the same state-action pair $(s, a)$ over and over. 
The agent also needs to explore other pairs.
This structure is similar to that of the bandit problem.

Therefore, by regarding each state-action pair $(s, a)$ as a slot machine and the change in $Q(s, a)$ as the reward in the context of bandit problems for the discontinuous Q-learning algorithm, we can associate the agent's attempt to select the state-action pair $(s, a)$ with large $\Delta Q$ as ``exploitation'' and the investigation of $\Delta Q$ for other state-action pairs $(s, a)$ as ``exploration.''
We thus define discontinuous bandit Q-learning (DBQL) as an algorithm that follows discontinuous Q-learning in which the bandit algorithm functions as the selection criterion.

In DBQL, the agent follows the softmax algorithm, a widely used algorithm to successfully solve the bandit problem.
The agent records $\Delta Q$ for each state-action pair $(s, a)$.
Let $\mu_t (s, a)$ be the empirical mean of $\Delta Q(s, a)$ by time step $t$.
The probability of the agent selecting the state-action pair $(s_i, a_j)$ at the next time step $t+1$ is calculated as follows:
\begin{equation}\label{eq:selec_prob}
    p_{t+1}(s_i, a_j) = \frac{e^{\beta \cdot \mu_t(s_i, a_j)}}{\sum\limits_{(s, a)} e^{\beta \cdot \mu_t(s, a)}},
\end{equation}
where $\beta$ controls the degree of exploration and exploitation.

\subsection{Multi-agent Learning}\label{subsec:multi_agent}
\begin{figure}[htp]
    \centering
    \includegraphics[width=0.9\textwidth]{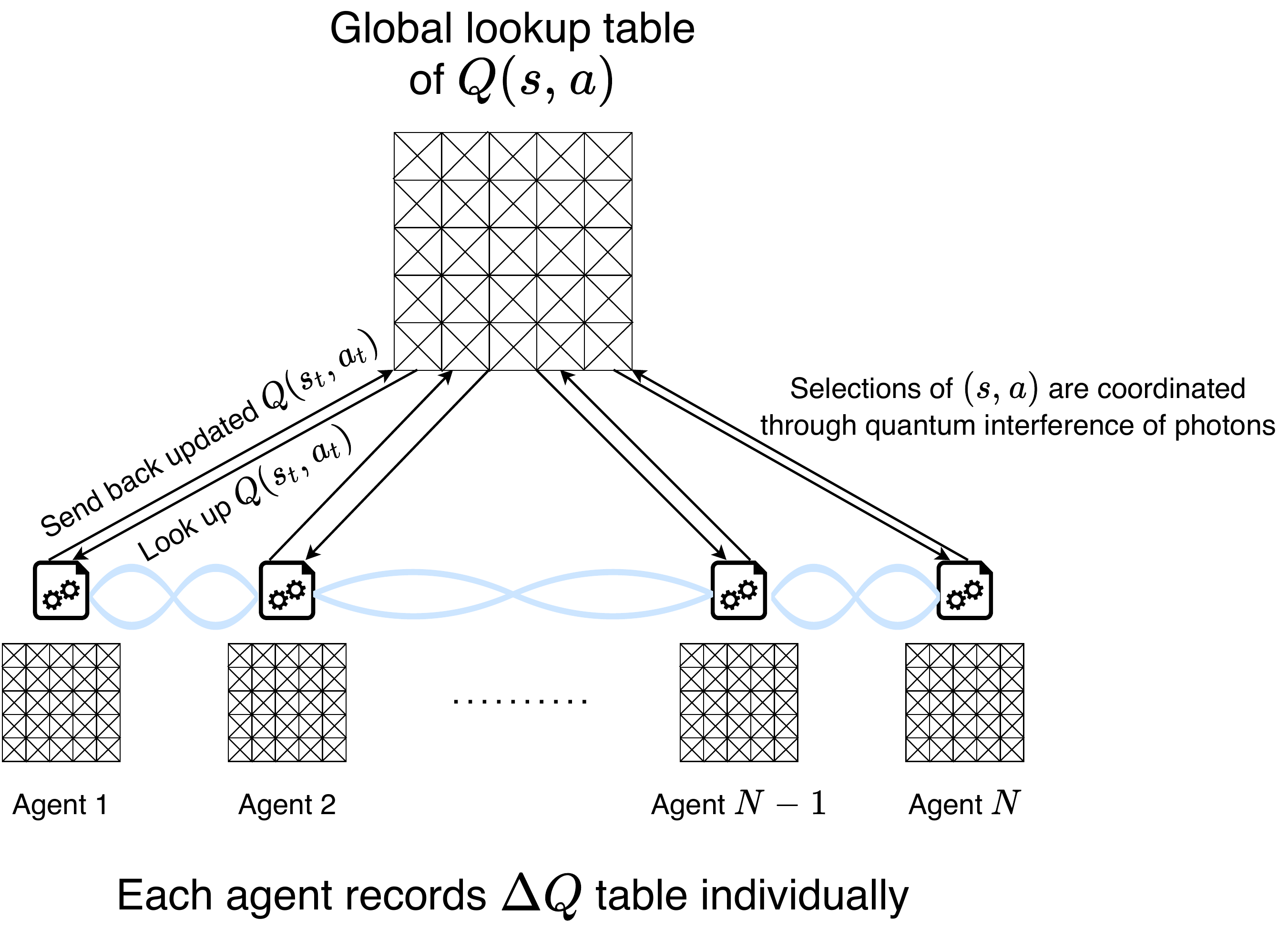}
    \caption{Structure of the DBQL by multiple agents. Each agent looks up a Q-value from the global lookup table, updates it using the generated reward and the next state provided by the environment, and sends it back to the global table. In our scheme, agents are not directly connected and cannot communicate with one another; yet, their state-action selections are coordinated owing to the quantum interference of photons. However, we coordinated their selections in this study using an algorithm because we could not extend the photonic system with two agents.}
    \label{fig:multi_dbql}
\end{figure}

In this study, multiple agents participate in simultaneously updating the global lookup table of $Q(s, a)$ based on DBQL to accelerate the learning process as shown in Fig. \ref{fig:multi_dbql}.
A situation is considered in which the agents share the global lookup table of $Q(s, a)$ while individually recording a separate table of $\Delta Q(s, a)$.
At time step $t$, each agent refers to the $\Delta Q$ table it has recorded and determines the state-action pair $(s_t, a_t)$ to update based on the softmax algorithm in Eq. \eqref{eq:selec_prob}.
Next, it retrieves the value of $Q(s_t, a_t)$ from the global lookup table, calculates the updated value of $Q(s_t, a_t)$ according to Eq. \eqref{eq:q_update}, and sends it back to the global table.

An important rule is that when two or more agents attempt to update the same state-action pair $(s, a)$ at the same time step, only one of the updates is randomly reflected to the global lookup table.
This depends on the problem settings; however, real-world examples exist in which the same investigation of state-action pairs by multiple agents is detrimental.
For example, consider an exploratory scenario where sonic waves are used to reveal the underlying stratigraphy of the seafloor.
Multiple agents simultaneously conducting the same location exploration would result in interference and yield poor results.

Moreover, even if we were to allow simultaneous updates by multiple agents and calculate the sum of $\Delta Q$ to reflect to the global table, this could disturb the convergence of Q-learning, because taking the sum essentially means changing the learning rate $\alpha$ locally for this particular time step.

\subsection{Cooperative Decision-Making through Quantum Interference}
\label{subsec:avoidance}
This section explains how the quantum interference of photons can be leveraged such that multiple agents can avoid selecting the same state-action pair $(s, a)$ at the same time step without any direct knowledge of the selections of the other agents.
As already mentioned, we have been unable to extend the conventional cooperative decision-making system with two agents, and thus the numerical demonstrations shown in Sec. \ref{sec:result} uses an algorithmic way to avoid selection conflicts.
Therefore, we will only cover the core concepts in this section and outline how in principle a photonic implementation may function.

Amakasu {\it{et al.}} \cite{amakasu2021conflict} proposed a conflict-free collective decision-making system with two agents using the orbital angular momentum (OAM) of light. 
A photon can carry theoretically an infinite number of OAMs, and the state of the photon is described as a superposition of different OAMs $|k \rangle$:
\begin{equation}
    |\Phi\rangle=\frac{1}{\sqrt{K}} \sum_{k=1}^{K} e^{i \phi_{k}}|+k\rangle.
\end{equation}
Because of the quantum property of photons, the detection probability of each OAM is calculated using the modulus square of the probability amplitude.
In addition, the usage of attenuators enables us to control the probability amplitudes, thus changing the observation probabilities.
In their proposed system, Amakasu {\it{et al.}} set $K$ equal to the number of options (in our case, this is the number of state-action pairs) and designed a protocol in which the agent selects the option whose index is the same as the detected OAM number.
For example, if an OAM of $|+1\rangle$ is detected by the first agent, the agent selects the first option.
This protocol enables probabilistic decision-making because the control of the probability amplitudes by the attenuators results in the control of the selection probabilities of the options.

\begin{figure}[htp]
    \centering
    \includegraphics[width=0.5\textwidth]{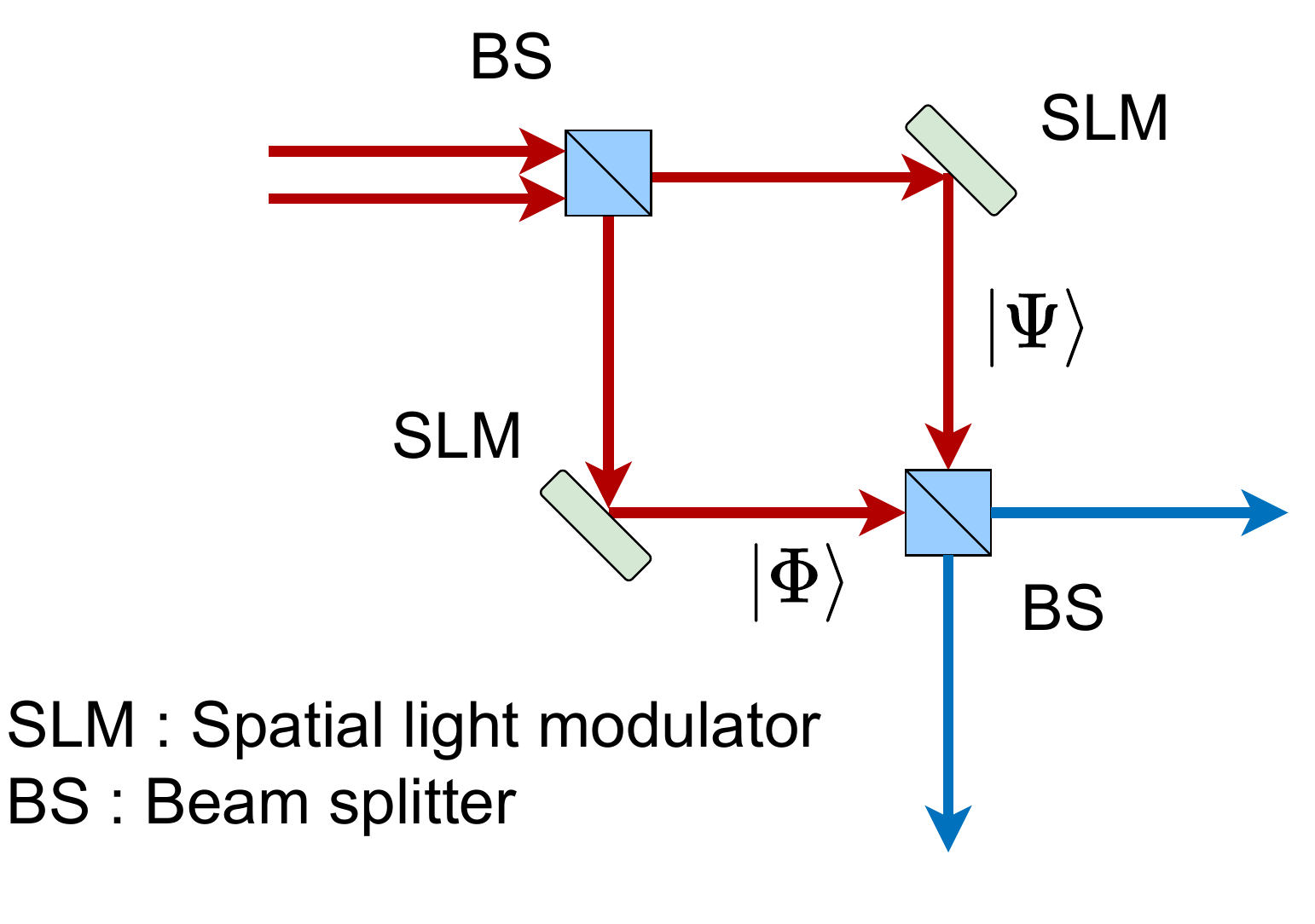}
    \caption{Two-photon Hong-Ou-Mandel interference. $|\Phi\rangle$ and $|\Psi\rangle$ represent the states of photons that are controlled by spatial light modulators.}
    \label{fig:hom}
\end{figure}

The use of quantum physics makes a difference when two agents simultaneously make decisions based on probability following the aforementioned protocol.
A quantum effect called the Hong-Ou-Mandel interference exists, whereby different OAMs are always observed when a photon-pair connected by this effect is observed by two detectors. 
Based on the protocol, the two agents always select different options without informing each other of their selections; that is, conflict-free selections are possible.
The implementation of the Hong-Ou-Mandel interference is quite simple and can be accomplished with only very basic optical instruments, such as spatial light modulators and beam splitters, as shown in Fig. \ref{fig:hom}.

\begin{figure}[htp]
    \centering
    \includegraphics[width=0.5\textwidth]{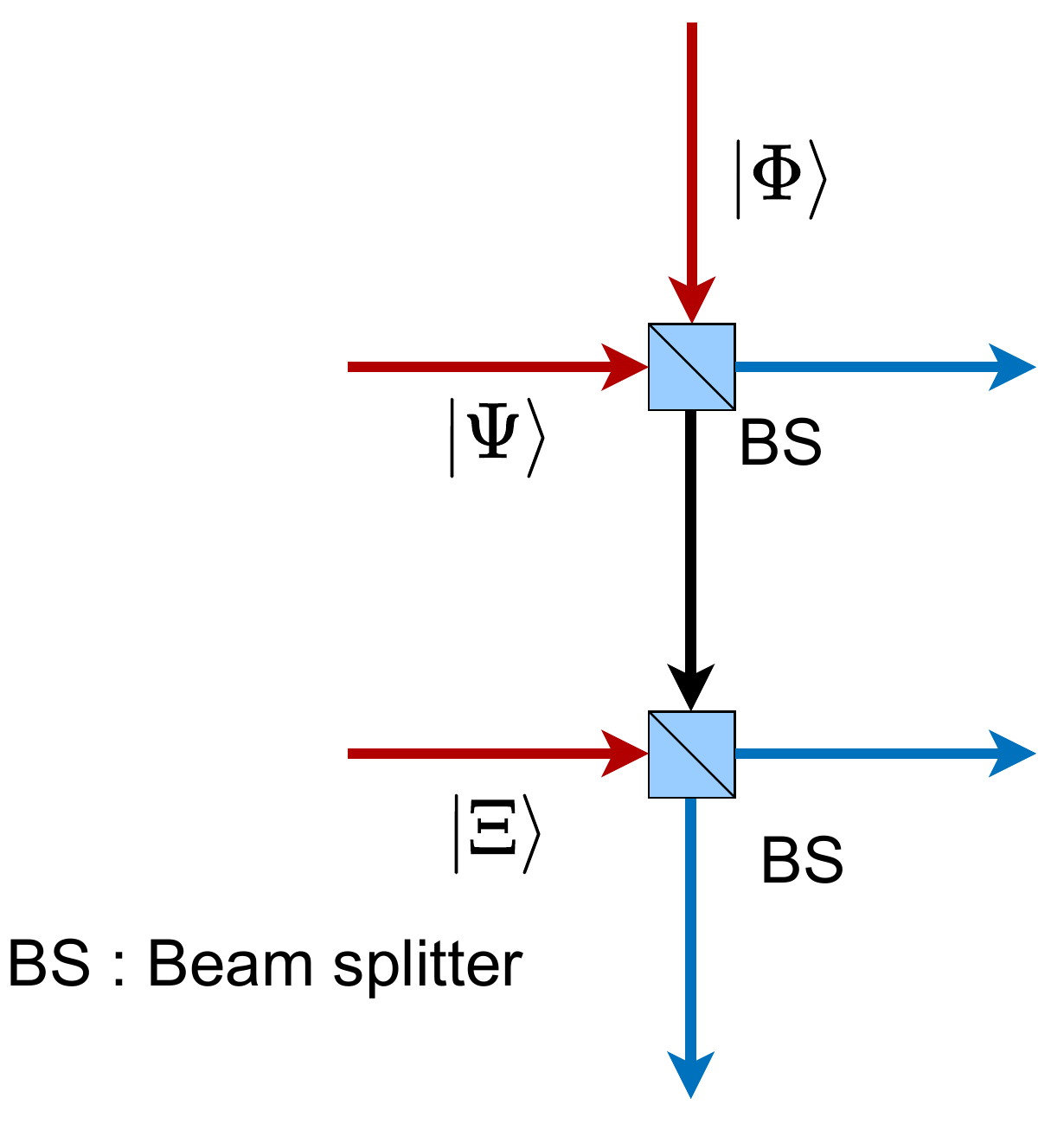}
    \caption{Photonic configuration with three photons. $|\Phi\rangle$, $|\Psi\rangle$, $|\Xi\rangle$ represent the states of photons.}
    \label{fig:three_photon_hom}
\end{figure}

Although a detailed design is yet to be devised, it is likely that conflict-free probabilistic decision-making can be realized with three or more agents by cascading multiple spatial light modulators and beam splitters as well as appropriately configuring the input OAM states as an extension of the previous setup.
For example, the schematic of a photonic configuration with three photons is shown in Fig. \ref{fig:three_photon_hom}.
This system eliminates selections completely in which all the agents select the same option; however, selections in which only two select the same option still remain.
Numerous studies on quantum interference among multiple photons have been conducted, including Refs. \cite{zukowski1997realizable, campos2000three, tillmann2015generalized}, thus successfully integrating these methods with the usage of OAMs has a guiding significance in developing a photonic system that completely eliminates selection conflicts in the future.

In our scheme, we assumed that the multi-photon conflict-free system can be realized and utilized to coordinate the probabilistic decision-making of $N$ agents through quantum interference of photons regarding the choice of the state-action pairs $(s, a)$.
This enables the agents to prevent selection conflicts without communicating with each other about the selection of pairs.
Not only is the learning accelerated because unnecessary updates are avoided, but also resources required to exchange information about state-action pair selections can be reduced.
Note that, simultaneous updates by different agents will be a waste except for one of them as explained in Sec. \ref{subsec:multi_agent}.
This study realized conflict-free selections using an algorithm by numerically computing the joint selection probabilities on a computer instead of using a photonic system.

\section{Results}\label{sec:result}
A $5\times 5$ grid world shown in Fig. \ref{fig:grid_world} was considered in this study, and we analyzed the situation in which multiple agents updates a state-action pair $(s, a)$ at every time step according to discontinuous Q-learning (Alg. \ref{alg:dql}).

\subsection{Rules in the Grid World}\label{subsec:rules}
The state-action combination in the grid world determines the rewards received by an agent and the cell to which it moves. The following settings are used in this study.
\begin{itemize}
    \item When any action is taken from cell A, the chances of the cell generating a reward of $+10$ is 50\% and the agent jumps to cell A'. If no reward is generated, it remains in cell A.
    \item When any action is taken from cell B, the chances of the cell generating a reward of $+5$ is 50\% and the agent jumps to cell B'. If no reward is generated, it remains in cell B.
    \item In any other cell, no reward is generated and the destination follows the action except when the agent hits a wall. In such cases, a reward of $-1$ is generated and the agent remains on the current cell.
\end{itemize}

\subsection{Objectives}
\begin{figure}[htp]
    \centering
    \includegraphics[width=0.5\textwidth]{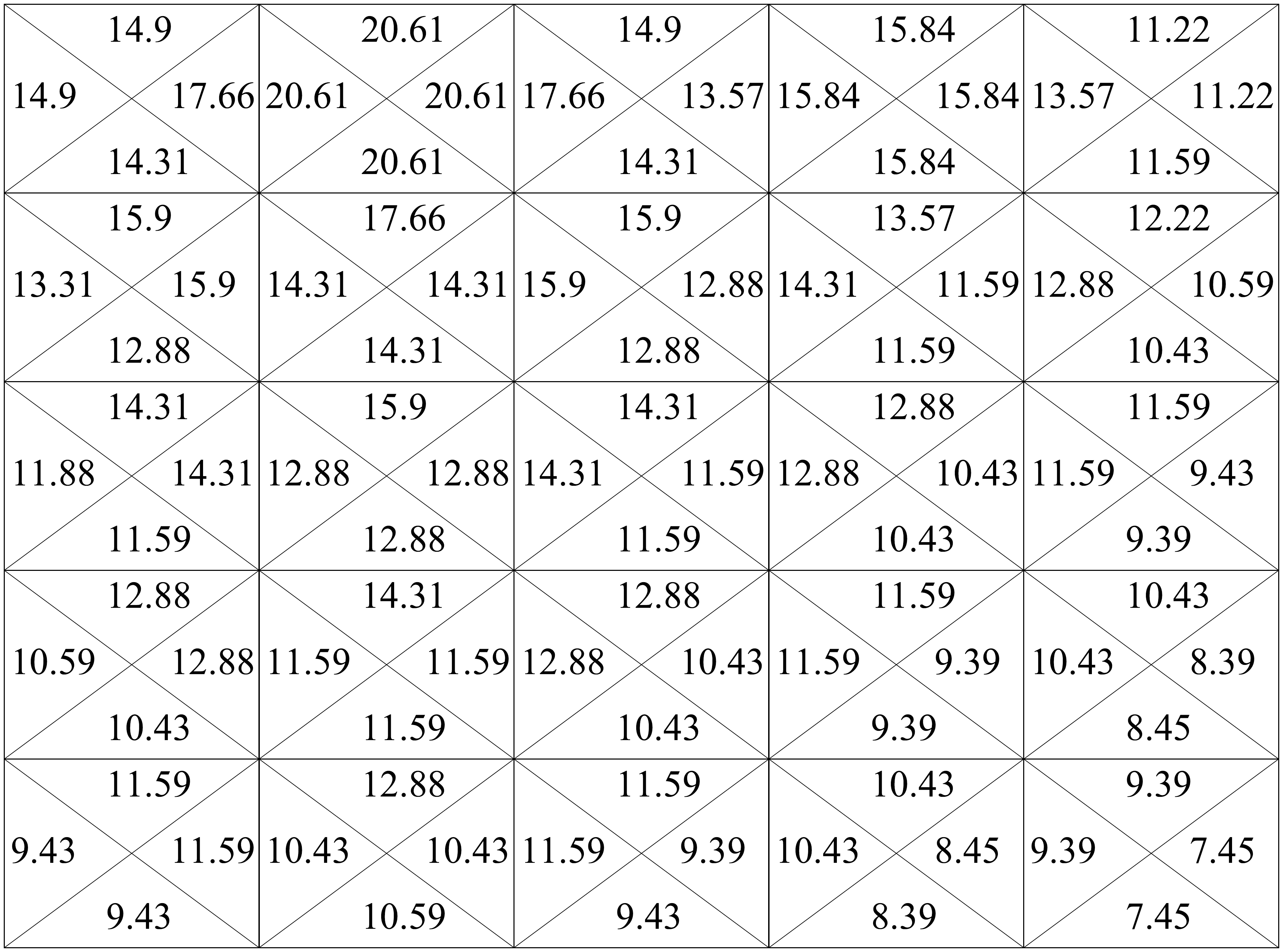}
    \caption{Ground truth values of the action-value function $\tilde{Q}(s, a)$.}
    \label{fig:true_q}
\end{figure}

Each agent selects a state-action pair $(s, a)$ at each time step and updates $Q(s, a)$ according to Alg. \ref{alg:dql}.
In this study, we considered 10--100 agents with 100 state-action pairs (25 cells and four actions in each cell).
To quantify the learning accuracy, we defined the loss $L_t$ as the average absolute error between the true action-value function $\tilde{Q}(s, a)$ and Q-values learned at time step $t$, which we refer to as $Q_t(s,a)$, over all the state-action pairs.
\begin{equation}
    L_t = \frac{1}{100}\sum_{(s, a)} |\tilde{Q}(s, a) - Q_t(s, a)|.
\end{equation}
One hundred trials were performed and the average loss $L_t$ over the trials was calculated as a metric to measure the gap between the optimal action-value functions and the Q-values.
The smaller the average loss, the more successful the learning process.
Based on the rules in Sec. \ref{subsec:rules}, the ground truth values of the action-value function $\tilde{Q}(s,a)$ are summarized in Fig. \ref{fig:true_q}.

Two major points were tested in this study: 
First, we tested whether the bandit algorithm outperforms random selections in Alg. \ref{alg:dql}.
That is, we compared the loss trajectories between DBQL and the case in which the agents make decisions uniformly at random instead of using the softmax algorithm as the selection criteria in Alg. \ref{alg:dql}.

Second, we considered the effect of conflict avoidance in the state-action pair selection on learning.
As noted in Sec. \ref{subsec:multi_agent}, if multiple agents simultaneously select the same state-action pair $(s, a)$, only one of their updates will be reflected in the global Q-table.
Hence, the learning process will always accelerate by avoiding selection conflicts.
This study aims to quantify this effect.
Furthermore, we demonstrate the significance of conflict avoidance especially when using DBQL.
The parameters are as follows: the number of iterations $T$ is 20000, learning rate $\alpha$ in Alg. \ref{alg:dql} is initially set to 0.035 that decays linearly to $\alpha = 0$ at $t = 20000$, and the time discount $\gamma$ is 0.9.
$\beta$ in Eq. \ref{eq:selec_prob}, which controls the degree of exploration and exploitation of the softmax algorithm used in DBQL, is initially set to 1.0 and grows linearly to $\beta = 5.0$ at $t = 20000$ because more exploitation is necessary in the later stage of learning.

\subsection{Performance Comparison}
\begin{figure}[htp]
    \centering
    \begin{subfigure}{0.9\textwidth}
        \centering
        \includegraphics[width=9.0cm]{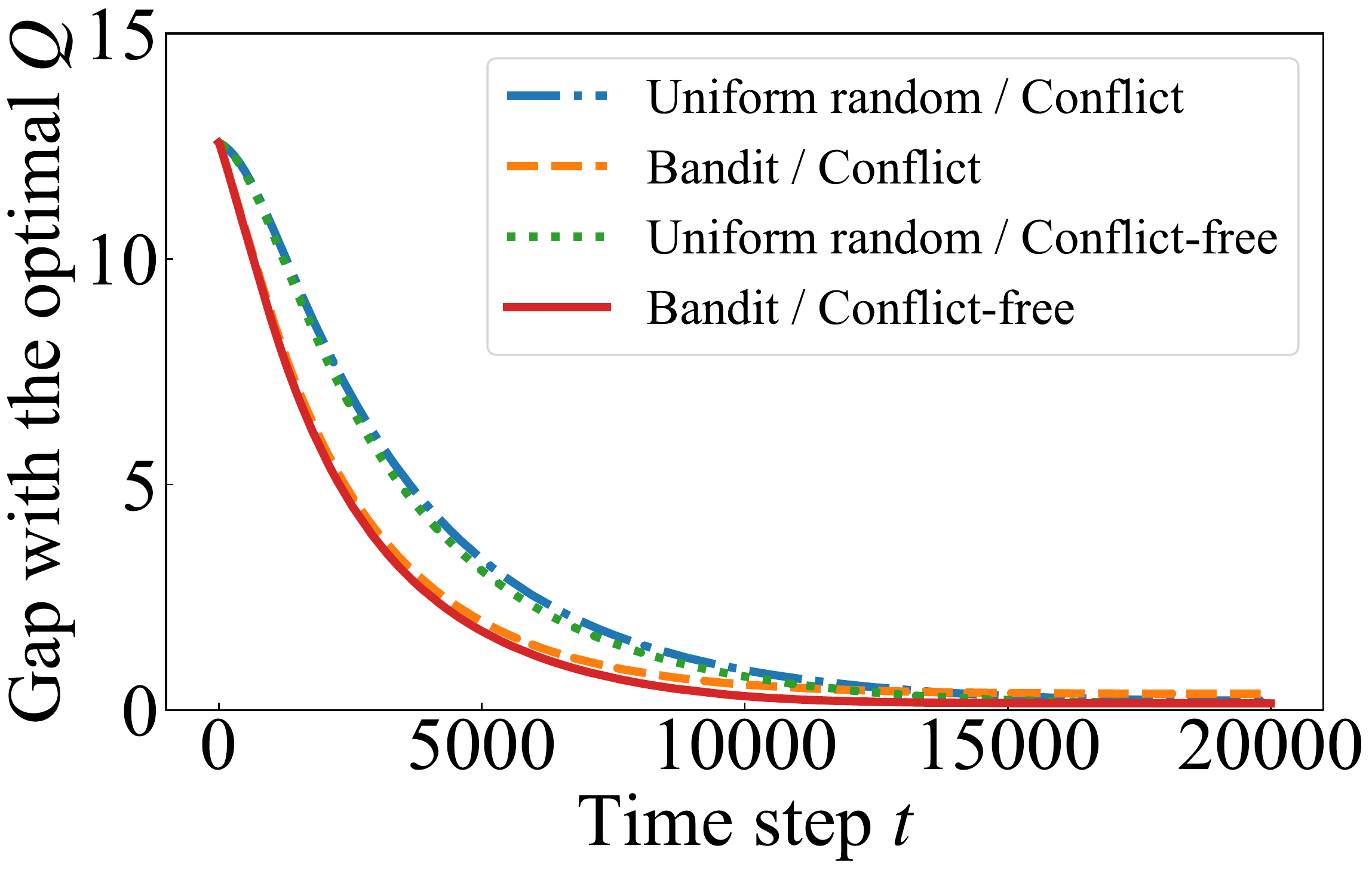}
        \caption{10 agents\label{fig:result_10agents}}
    \end{subfigure}\\
    \begin{subfigure}{0.9\textwidth}
        \centering
        \includegraphics[width=9.0cm]{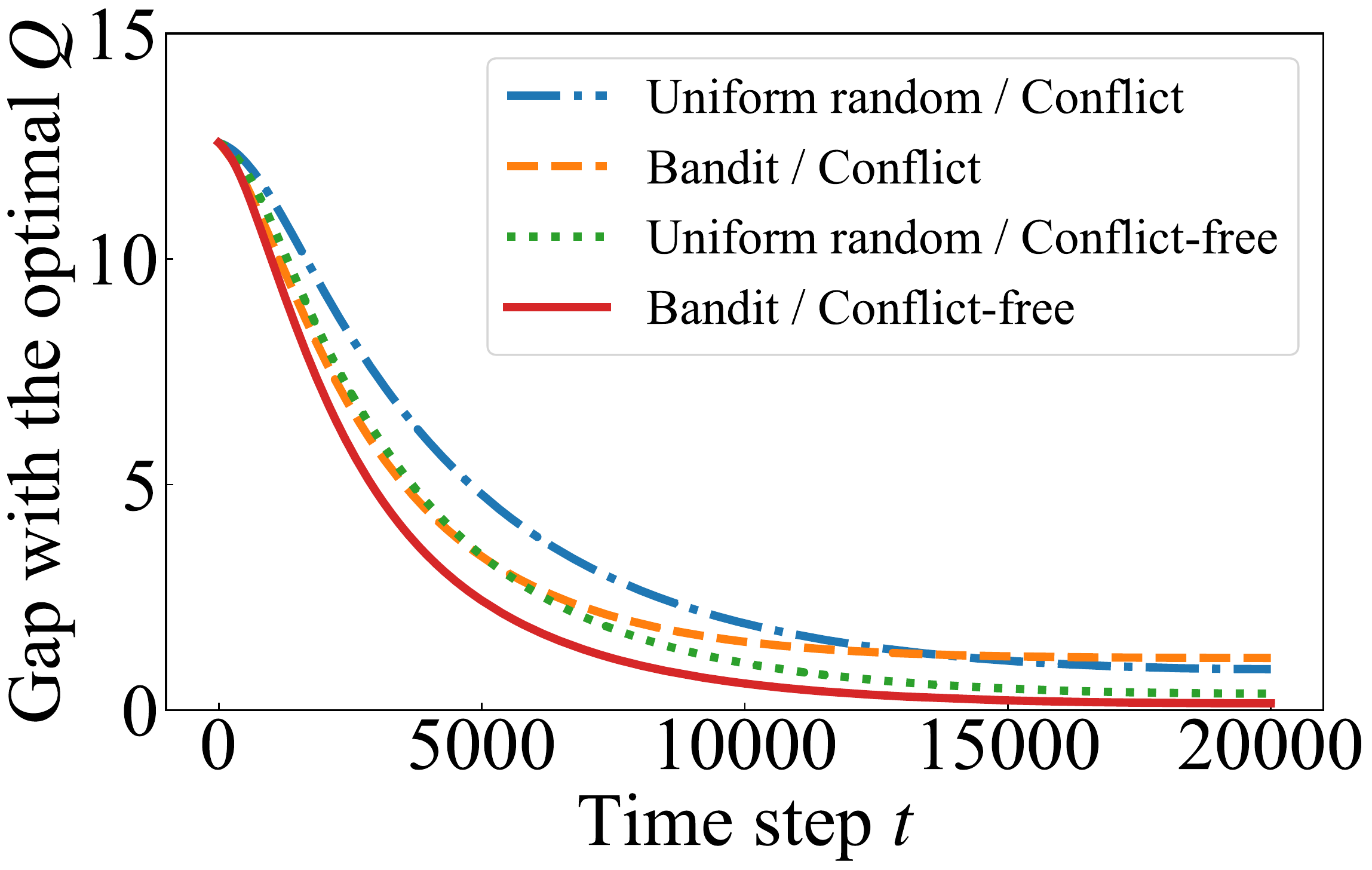}
        \caption{50 agents\label{fig:result_50agents}}
    \end{subfigure}\\
    \begin{subfigure}{0.9\textwidth}
        \centering
        \includegraphics[width=9.0cm]{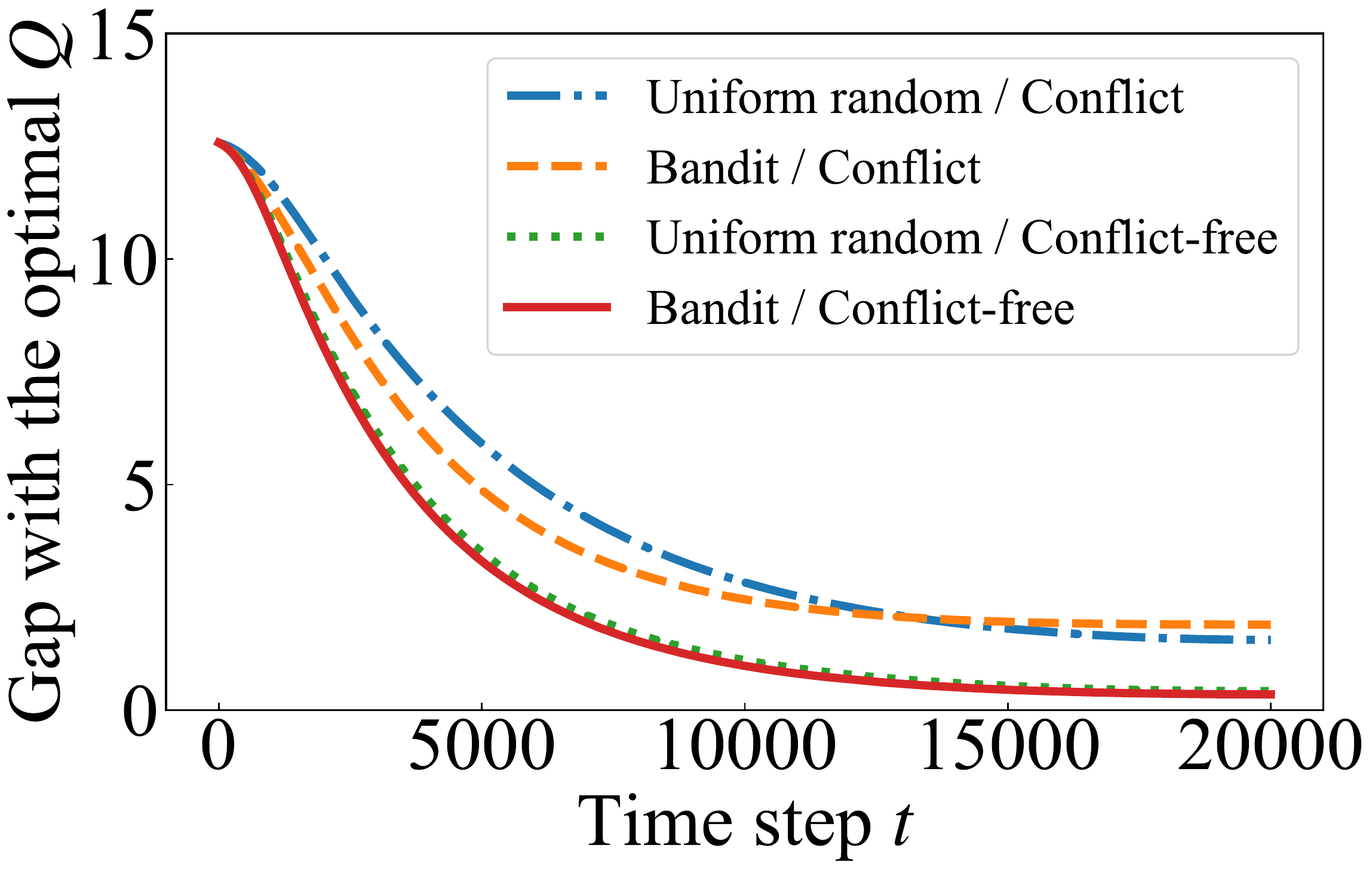}
        \caption{90 agents\label{fig:result_90agents}}
    \end{subfigure}
    \caption{Comparison of the four learning methods. The average loss, representing the gap between optimal action-value functions and the Q-values during learning, is shown for the four learning methods. The learning methods are divided along two axes: whether the bandit algorithm is used for selection criteria and whether selection conflicts are allowed among different agents.}
    \label{fig:result_loss}
\end{figure}

The average loss $L_t$, which quantifies the gap between the optimal action-value functions and the Q-values during learning when the number of agents is 10, 50, or 90, are shown in Figs. \ref{fig:result_loss} (a), (b), and (c), respectively.
Therein, the blue and orange curves represent the cases in which different agents were allowed to simultaneously select the same state-action pair, denoted by ``conflict.''
The procedure for state-action pair selections differs for the blue and orange curves. 
The blue curve denoted by the legend ``uniform random/conflict'' is based on random selections, whereas the orange curve denoted by ``bandit/conflict'' is based on the softmax algorithm in Eq. \eqref{eq:selec_prob}.

Similarly, the green and red lines represent the cases in which the state-action pair selections are conducted in a ``conflict-free'' manner, as marked by the latter half of the legend. 
Agents were not allowed to select the same state-action pair at the same time step.
In addition, actions were selected in a uniformly random manner in the green curve, whereas the red curve was based on a bandit-based approach. 
Hence, the green and red curves were denoted as ``uniform random/conflict-free'' and ``bandit/conflict-free,'' respectively.

First, we compared the random-based lines with the bandit-based lines to examine the effect of the bandit algorithm.
By comparing the blue and orange lines or the green and red lines, we observed that the learning is faster when the agents follow the bandit algorithm.
This validates the effectiveness of DBQL, which considers the change in the Q-value ($\Delta Q$) as the reward of the bandit problem.

As the number of agents approaches a hundred, the difference in performances between uniform random/conflict-free and bandit/conflict-free narrows.
This is because in situations where decision overlaps are not allowed, the significance of each agent's sensible selection is reduced if the number of agents is sufficiently large.
In particular, when the number of agents is one hundred, the two performances are exactly the same, irrespective of the choice made by the agents because only one agent is always assigned to each state-action pair.

Moreover, we analyzed the impact of conflict avoidance in state-action pair selections on learning.
While the results are rather obvious considering the environmental rules, learning is faster when the selections are conflict-free.
We defined $S_{\text{under}}$ as the area under the learning curve to quantify the learning efficiency across the entire learning process.
\begin{equation}
    S_{\text{under}} = \sum_{t = 1}^{T} L_t.
\end{equation}

Table \ref{tab:conflict_avoidance} lists the $S_{\text{under}}$ ratios of uniform random/conflict by uniform random/conflict-free and bandit/conflict by bandit/conflict-free to quantify the benefit of the conflict avoidance in state-action pair selections.

\begin{table}[htp]
    \caption{Benefit of conflict avoidance in the selection of state-action pairs. Conflict avoidance is crucial when the number of agents increases.} 
    \centering
    \begin{tabular}{|c|c|c|c|c|c|c|c|c|c|c|}
            \hline
            Number of agents & 10 & 20 & 30 & 40 & 50 & 60 & 70 & 80 & 90 & 100\\
            \hline
            Uniform random & 1.06 & 1.12 & 1.19 & 1.26 & 1.32 & 1.39 & 1.46 & 1.53 & 1.59 & 1.67 \\  
            \hline
            Bandit & 1.13 & 1.26 & 1.34 & 1.4 & 1.43 & 1.47 & 1.49 & 1.52 & 1.55 & 1.56\\
            \hline
    \end{tabular}
    \label{tab:conflict_avoidance}
\end{table}

As the number of agents increases, the ratio also increases, indicating that conflict avoidance provides a more significant benefit.
This is because the probability of multiple agents selecting the same state-action pair $(s, a)$ increases and more selections are discarded when selection conflicts are allowed, as only one of them is valid.

\begin{figure}[htp]
    \centering
    \includegraphics[width=0.5\textwidth]{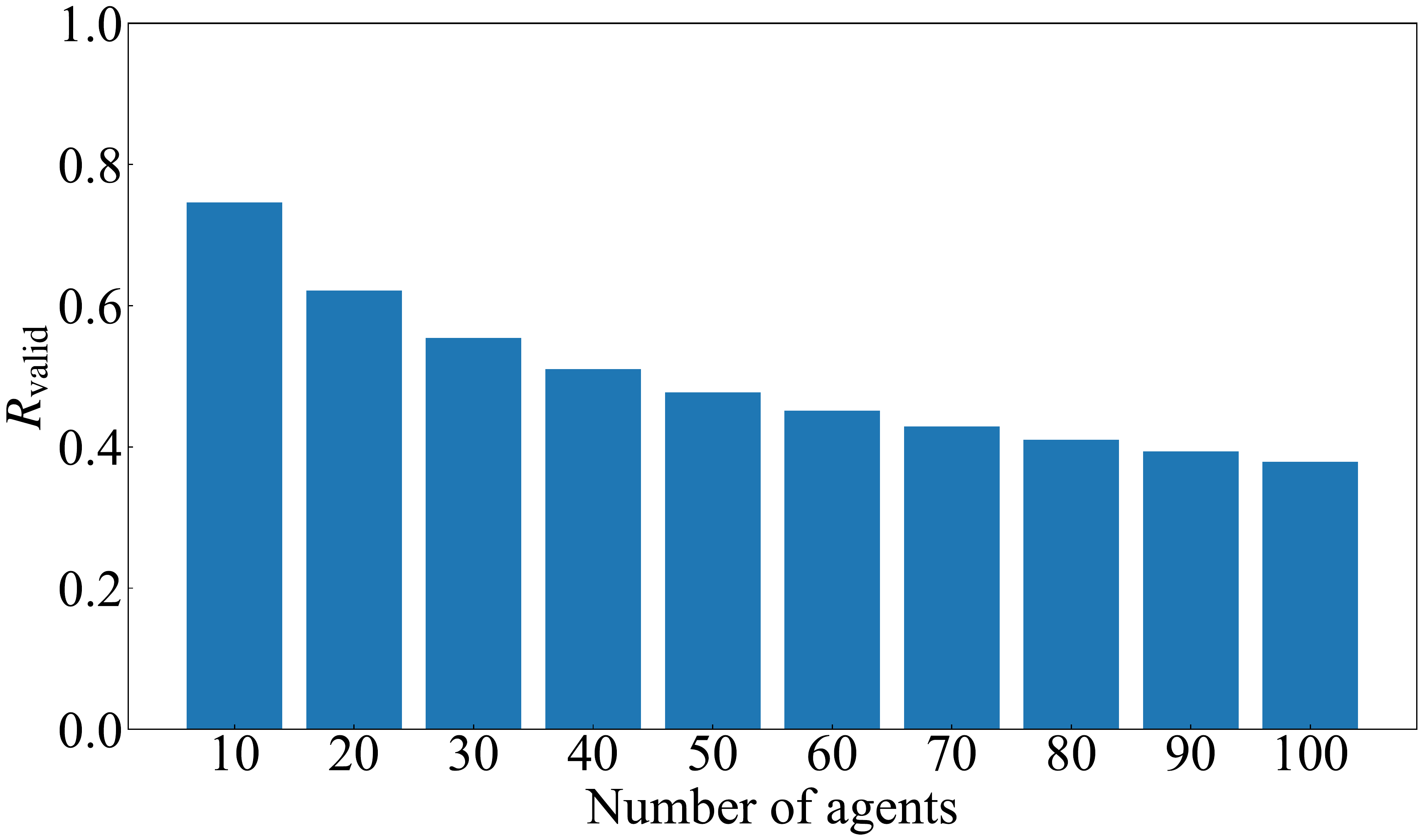}
    \caption{Valid selection rate $R_{\text{valid}}$. The larger the number of agents, the more frequently the selections of the agents overlap, validating the significance of conflict avoidance.}    \label{fig:valid_choice}  
\end{figure}

Thus, we defined $R_{\text{valid}}$ as the proportion of valid choices to quantitatively evaluate such effects.
The change in $R_{\text{valid}}$ for bandit/conflict as the number of agents changes is shown in Fig. \ref{fig:valid_choice}.
 $R_{\text{valid}}$ decreased as the number of agents increased, indicating that conflict avoidance is more significant for an increased number of agents.
When the number of agents is one hundred, approximately 60\% of the updates are wasted if the agents' selections are not coordinated.

\begin{figure}[htp]
    \centering
    \includegraphics[width=0.5\textwidth]{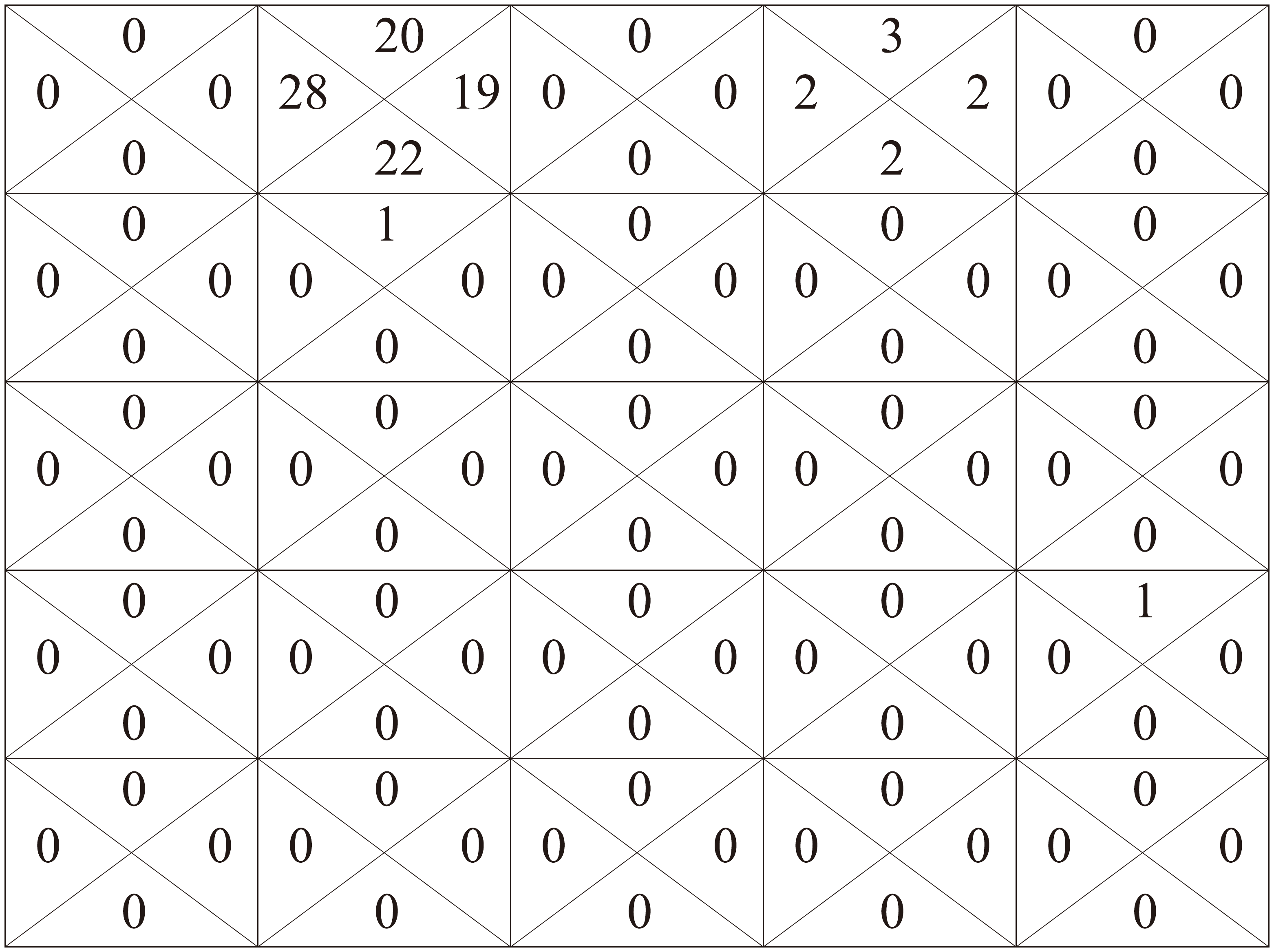}
    \caption{Number of agents that choose each state-action pair $(s, a)$ at the final time step for the bandit-based algorithm when conflicts are allowed. Approximately 90\% of the agents update cell A, which generated the highest reward, while the other cells remain largely unexplored.}
    \label{fig:agent_choices}  
\end{figure}

Furthermore, comparing the convergence values in Fig. \ref{fig:result_loss}, bandit/conflict has a larger value than random/conflict, indicating that uniform random/conflict outperforms bandit/conflict if only the final accuracy is compared.
This is because, in bandit/conflict, most agents decide to update cell A as they proceed with learning; therefore, other cells are not updated, thus resulting in residual action-value function errors for those cells.
Figure \ref{fig:agent_choices} shows an example of the number of agents that choose each state-action pair $(s, a)$ in the final time step when the number of agents is a hundred for the bandit/conflict case.
Approximately 90\% of the agents update cell A and most of the remaining agents update cell B, which has the second highest reward.
Therefore, ensuring the conflict-free property is crucial to avoiding most agents getting ``stuck'' when using a bandit-derived algorithm.

\section{Discussion}\label{sec:discussion}
This study proposed a photonic reinforcement learning scheme, which required both a novel algorithm and photonic system, and demonstrated its performance.
We employed the grid world problem, a frequently used model in reinforcement learning, with the aim to learn the optimal action-value functions for all state-action pairs $(s, a)$ with high precision, involving multiple agents. 
The details presented in the study are summarized as follows.

First, we proposed DBQL, a learning algorithm in which each agent selected one of all the state-action pairs $(s, a)$ in the environment in each time step $t$ and updated $Q(s, a)$ based on the same formula as in the original Q-learning.
The decision-making problem of selecting a state-action pair $(s, a)$ was similar to the bandit problem.
This is because if we define the amount of $Q(s, a)$-update at each time step as $\Delta Q$, the agent must strike a balance between the demand to select state-action pairs $(s, a)$ with large $\Delta Q$ to accelerate the learning process (``exploitation'' in the context of the bandit problem) and the demand to investigate the values of $\Delta Q$ for other state-action pairs $(s, a)$ (``exploration'' in the bandit problem).
We compared the case in which the bandit algorithm was used as the decision-making criteria for Alg. \ref{alg:dql} with that in which a uniform random selection was used, to validate the effectiveness of DBQL.
The former resulted in a faster learning process as described in Sec. \ref{sec:result}.

Second, we proposed a multi-agent architecture where multiple agents make conflict-free decisions in the learning, and then quantitatively evaluated the impact of the conflict avoidance on learning.
As demonstrated in Sec. \ref{sec:result}, the learning was indeed accelerated by avoiding selection conflicts particularly when the number of agents increased.
Moreover, cooperative decision-making is essential when multiple agents follow DBQL to avoid getting stuck in the later stage of learning.
Without coordinating the selection, most agents would be highly likely to select the cell with largest reward in the latter parts of the learning.
Although the particular configuration of the system is not yet established, a photonic system with cascaded spatial light modulators and beam splitters is expected to enable cooperative decision-making by three or more agents for avoiding selection conflicts.
Once this system is conceived in the future, it can be incorporated into our proposed scheme and obviate the necessity for the agents to communicate with each other to coordinate their selections.

While Amakasu {\it{et al.}} provided the concept of the base idea that addressed the competitive bandit problem, this study addresses a general reinforcement learning problem with the grid world problem as an example.
The two problems differ in that, in the bandit problem, the machines' reward probabilities are invariant regardless of the agent's action; however, in the grid world problem, state transitions, which correspond to the changes in the reward probabilities in the context of the bandit problem, occur because of the agent's action.
Our proposed scheme applies to such challenging problems in a dynamic environment.

Next, we discuss some of the limitations of this study and how they may be addressed in the future.
First, in DBQL, the agent's actions are discontinuous. 
This can be resolved by restricting the possible state-action pairs $(s, a)$ that can be chosen by the agent at each time step to those in the current cell.
However, with this method, if more than four agents end up in a particular cell, at least two will have to choose the same state-action pair $(s, a)$ in the next time step.
This requires rule making for exception handling.
Second, when the number of state-action pairs is sufficiently larger than the number of agents, conflicts of choice occur less frequently, and the advantage of conflict avoidance by quantum interference may be reduced.
Regarding this concern, as indicated in the birthday paradox, the probability that the choices of two agents overlap is greater than our intuition even if the number of agents is such smaller than that of pairs. 
For example, suppose 100 state-action pairs exist and 10 agents are to make uniform choices at random, the probability that at least two agents make the same choice is over 37\%.
Furthermore, as mentioned in Sec. \ref{sec:result}, conflict avoidance is essential in DBQL because the probability of multiple agents intending to make the same choice increases significantly as learning proceeds.

In the future, our first priority is to design a system that allows conflict-free decision-making by three or more agents.
Additionally, we would like to develop algorithms that allow agents to take continuous actions and apply DBQL to other reinforcement learning problems that are more complex than the grid world.
To the best of our knowledge, this study is the first to connect the notion of photonic cooperative decision-making with Q-learning and apply it to a dynamic environment. We believe this study makes a valuable contribution to the field of decision-making using physical processes.

\subsection*{Data Availability}
Data used in this study are available from the corresponding author upon request.

\subsection*{Conflicts of Interest}
The authors declare that there are no conflicts of interest regarding the publication of this paper.

\subsection*{Acknowledgments}
This study was supported in part by the CREST project (JPMJCR17N2) funded by the Japan Science and Technology Agency, Grants-in-Aid for Scientific Research (JP20H00233) and Transformative Research Areas (A) (JP22H05197) funded by the Japan Society for the Promotion of Science. AR was funded by the Japan Society for the Promotion of Science as an International Research Fellow.

\printbibliography

\end{document}